# Low-discrepancy Sampling in the Expanded Dimensional Space: An Acceleration Technique for Particle Swarm Optimization

Feng Wu*, Yuelin Zhao, Jianhua Pang*, Jun Yan, and Wanxie Zhong

*State Key Laboratory of Structural Analysis, Optimization and CAE Software for Industrial Equipment, Department of Engineering Mechanics, Faculty of Vehicle Engineering and Mechanics, Dalian University of Technology, Dalian 116023, P.R.China; Guangdong Ocean University, Zhanjiang 524088, China; Shenzhen Institute of Guangdong Ocean University, Shenzhen 518120, China*

*Email:, wufeng_chn@163.com, zhaoyl9811@163.com, pangjianhua@gdou.edu.cn, yanjun@dlut.edu.cn, wxzhong@dlut.edu.cn*

Project supported by the National Natural Science Foundation of China grants (Nos. 11472076, 51609034 and U1906233), the National Key R&D program of China (No. 2021YFA1003501), the Science Foundation of Liaoning Province of China (No. 2021-MS-119), the Dalian Youth Science and Technology Star project (No. 2018RQ06), and the Fundamental Research Funds for the Central Universities grant (Nos. DUT20RC(5)009 and DUT20GJ216).

*Corresponding author: wufeng_chn@163.com; *pangjianhua@gdou.edu.cn*

March 2023

**Abstract:** Compared with random sampling, low-discrepancy sampling is more effective in covering the search space. However, the existing research cannot definitely state whether the impact of a low-discrepancy sample on particle swarm optimization (PSO) is positive or negative. Using Niderreiter's theorem, this study completes an error analysis of PSO, which reveals that the error bound of PSO at each iteration depends on the dispersion of the sample set in an expanded dimensional space. Based on this error analysis, an acceleration technique for PSO-type algorithms is proposed with low-discrepancy sampling in the expanded dimensional space. The acceleration technique can generate a low-discrepancy sample set with a smaller dispersion, compared with a random sampling, in the expanded dimensional space; it also reduces the error at each iteration, and hence improves the convergence speed. The acceleration technique is combined with the standard PSO and the comprehensive learning particle swarm optimization, and the performance of the improved algorithm is compared with the original algorithm. The experimental results show that the two improved algorithms have significantly faster convergence speed under the same accuracy requirement.

# 1. Introduction

article swarm optimization (PSO) is a population-based stochastic optimization technology. Since it was proposed by Kennedy and Eberhart in 1995 [1, 2], it has attracted great attention. PSO is an excellent evolutionary algorithm that is easy to implement, can quickly converge to the optimal solution, and has been proven to perform well in many fields, such as neural networks [3, 4], photovoltaic system [5], feature selection [6, 7], operational research [8], etc.

However, it is not easy for PSO to consider both the convergence speed and calculation accuracy when solving complex optimization problems. Many excellent PSO variants are committed to solving this problem, and can be divided based on three aspects. The first aspect is to enhance the PSO by adjusting the parameter changes in the algorithm; Shi and Eberhart [9] introduced the inertia weight, which can balance the local development and global search capabilities of PSO. Clerc and Kennedy [10] proposed a set of coefficients to control the convergence trend of the system. Zhan *et al.* [11] developed an adaptive PSO and improved the search efficiency. There are also a sea of enhancement approaches for the PSO iteration formula. Liang *et al.* [12] proposed a new comprehensive learning particle swarm optimizer (CLPSO), which introduces a learning strategy to update the particle's velocity that significantly improves the performance of PSO. Lynn and Suganthan [13] proposed heterogeneous comprehensive learning particle swarm optimization (HCLPSO), which divides the population into two subpopulations, and each subpopulation only focuses on exploration or exploitation to maintain the diversity of the population. Zhan *et al.* [14] proposed orthogonal learning particle swarm optimization, which employs an orthogonal learning strategy that can guide particles to fly in an optimal direction. Liu et al. [15] proposed an improved quantum particle swarm optimization with Lévy flight and straight flight, which is more suitable for high-dimensional situations. Mendes *et al.* [16] and Zeng *et al.* [17] showed that replacing the best position among all the particles with the best position in some smaller number of adjacent particles can also improve the search ability of PSO. The third aspect focuses on population initialization [18, 19]. In general, the population is generated by random samples [20, 21]. However, quite a few researchers believe that the uniformity of the initial population obviously impacts the search ability of PSO. Employing numerical experiments, Richards and Ventura [22] concluded that low-discrepancy sampling (LDS) can be used to improve the calculation accuracy and the convergence speed of PSO in high-dimensional space. Morrison [23] verified, based on numerical simulations, that the LDS initialization method can reduce the variation of the search results without losing accuracy and performance. Nevertheless, some studies still disagree with this view. Li *et al.* [24] showed through the numerical results that random initialization is most suitable for PSO. Tharwat and Schenck [25] compared the effects of five famous population initialization methods and concluded that PSO is not sensitive to initialization methods under sufficient numbers of iterations.

It can be said that the existing studies did not give a clear conclusion on whether the uniformity of the initial sample set has a positive or negative impact on the PSO algorithm and why this impact occurs. The author has also done experiments, and the result show that PSO is not sensitive to initialization methods. The author has further verified the effect of using LDS directly in each iteration, and the result is worse. However, it is undeniable that many experiments observe that in the early iterations process, the population generated by LDS has a better convergence speed. Then, why can't the advantages of using an LDS initialization at the early iterations be preserved during the entire iteration of PSO? Furthermore, what is the relationship between the error of the entire iteration process of PSO and the sample uniformity? These unclear conclusions and contradictions motivated us to investigate more deeply. First, we try to find the answers to these two questions from the existing theoretical analysis.

Currently, the theoretical study of the PSO algorithm mainly focuses on two aspects. One is the movement of the single particle or the particle swarm with time; Kennedy [26] simulated different particle trajectories by simplifying the particle evolution model. Ozcan and Mohan [27] pointed out that particles move along a path defined by a sinusoidal wave in a simplified one-dimensional PSO system. van den Bergh [28] developed a model that could describe the behavior of a single particle to investigate the local convergence properties of PSO. Liang *et al.* [12] analyzed the particle swarm system's search behavior and concluded that the CLPSO algorithm can search a greater number of potential regions. Another aspect of theoretical analysis is studying PSO's stability and convergence. Clerc and Kennedy [10] first analyzed the stability of PSO. Trelea [29], Emara and Fattah [30] determined the parameter range that can ensure the stability or convergence of the algorithm through a similar analysis. García-Gonzalo and Fernández-Martínez [31] provided an analytical expression for the upper bound of the second-order stable regions of the particle trajectories. Kadirkamanathan *et al.* [32] analyzed the stability of the particle dynamics under stochastic conditions using a Lyapunov stability analysis and the concept of passive systems.

From the current work, it is easy to see that although many studies have focused on a theoretical analysis of the PSO algorithm, an error analysis of PSO urgently needs research attention. Moreover, the existing theoretical analysis cannot explain the two questions raised in the previous article. Therefore, this study attempts to answer these two questions. Using Niderreiter's theorem, this study analyzes the error of PSO and proposes a new acceleration technique. Some representative LDSs are selected for numerical validation. At present, there are many types of LDSs. For example, there are sampling methods based on orthogonal experimental design, including the Latin hypercube [33], orthogonal array [34], uniform experimental design [35] and so on. There are also sampling methods based on number theory, which often have relatively simple generation formats, such as the Hua-Wang sampling (HWS) [36-38], Halton sampling (HS) [39], Sobol sampling (SS) [40] and so on. In addition, there are some other sampling methods, such as Chaos sampling [41], Poisson disk sampling [42, 43], optimized Halton sampling (OHS) [44], and dynamics evolution sampling (DES) [45]. The OHS uses an evolutionary algorithm to optimize HS, and the DES uses a dynamic evolutionary algorithm to optimize the existing samples. Both methods generate a large number of samples, which can be downloaded directly on the internet. We choose HWS, DES, SS and Halton scramble sampling (HSS) [46], a modified version of the famous HS for numerical verification. The research contributions of this study are as follows:

1) Based on Niderreiter's theorem, an error analysis of PSO is given, and the influence of LDS initialization on the PSO calculation results is clarified. The relationship among sample uniformity, calculation error and the convergence rate is found, which provides a new theoretical basis for improving PSO.

2) Based on the theoretical derivation in 1), a new acceleration technique, LDS in the expanded dimensional space (LDSEDS), is given. According to different sample construction methods, two versions of LDSEDS are proposed. These two versions are combined with PSO to generate two more effective algorithms. In the simulation studies, several benchmark functions are performed, and the performances of the proposed algorithms are compared with the original PSO to demonstrate the superiority of these two algorithms. At the same time, the computational performance of these two algorithms is compared, and a more effective algorithm is selected for higher dimension promotion.

3) The LDSEDS acceleration technique proposed in 2) continues to be extended to other PSO series algorithms. Considering that the theoretical derivation in 1) needs preconditions, CLPSO is ultimately selected for an improvement to form CLPSO-LDSEDS, and the superiority of CLPSO-LDSEDS is also verified.

The rest of this paper is organized as follows. Section 2 briefly introduces the preliminaries. Section 3 gives the error analysis of PSO. Section 4 proposes a new acceleration technique LDSEDS to form two acceleration algorithms. In Section 5, the acceleration algorithms are verified numerically. In Section 6, this acceleration technique is extended to CLPSO. Finally, Section 7 draws a conclusion and gives future work directions.

# 2. Preliminaries

## 2.1 Particle Swarm Optimization

Without loss of generality, we assume a $D$- dimensional optimization problem to be expressed as

$$\begin{cases} \min\ f(\boldsymbol{x}) = f(x_1, x_2, \cdots, x_D) \\ a_i \le x_i \le b_i \end{cases}. \tag{1}$$

where $f(\boldsymbol{x})$ is the cost function, and $a_i$ and $b_i$ represent the boundary of the search space. To solve Eq.(1), we use PSO, which simulates the swarming habits of animals such as birds [1]. The positions and velocities of all the particles at the $g$-th iteration are expressed as

$$\boldsymbol{X}_g = [\boldsymbol{x}_{g,1}, \boldsymbol{x}_{g,2}, \cdots, \boldsymbol{x}_{g,N}],\ \boldsymbol{V}_g = [\boldsymbol{v}_{g,1}, \boldsymbol{v}_{g,2}, \cdots, \boldsymbol{v}_{g,N}], \tag{2}$$

where $N$ is the number of particles. The position and velocity of the $i$-th particle are updated using

$$\begin{aligned} &\boldsymbol{v}_{g+1,i} = \omega \boldsymbol{v}_{g,i} + c_1 \boldsymbol{\varepsilon}_{g,i}^{(1)} \circ \left(\boldsymbol{p}_{g,i}^{(l)} - \boldsymbol{x}_{g,i}\right) + c_2 \boldsymbol{\varepsilon}_{g,i}^{(2)} \circ \left(\boldsymbol{p}^{(g)} - \boldsymbol{x}_{g,i}\right) \\ &\boldsymbol{x}_{g+1,i} = \boldsymbol{x}_{g,i} + \boldsymbol{v}_{g+1,i} \end{aligned}, \tag{3}$$

where $\omega$, $c_1$ and $c_2$ are coefficients that can be constants or change with iteration, $\circ$ denotes the Hadamard product, $\boldsymbol{\varepsilon}_{g,i}^{(1)}$ and $\boldsymbol{\varepsilon}_{g,i}^{(2)}$ are two $D \times 1$ random vectors of numbers distributed between 0 and 1, and

$$\boldsymbol{p}^{(g)} = \arg \min_{1 \le i \le N, 1 \le l \le g} f(\boldsymbol{x}_{l,i}), \quad \boldsymbol{p}_{g,i}^{(l)} = \arg \min_{1 \le l \le g} f(\boldsymbol{x}_{l,i}). \tag{4}$$

## 2.2 Comprehensive Learning Particle Swarm Optimizer

CLPSO is a variant of PSO that uses a new learning strategy to update the particle's velocity [12]. The position and velocity in CLPSO are updated using

$$\boldsymbol{v}_{g+1,i} = \omega \boldsymbol{v}_{g,i} + c \boldsymbol{\varepsilon}_{g,i} \circ (\boldsymbol{p}_{g,ij} - \boldsymbol{x}_{g,i}), \quad \boldsymbol{x}_{g+1,i} = \boldsymbol{x}_{g,i} + \boldsymbol{v}_{g+1,i}, \tag{5}$$

where $\omega$ and $c$ are coefficients that can change with the iterations. $\boldsymbol{\varepsilon}_{g,i}$ is a $D \times 1$ random vector of numbers distributed between 0 and 1. The $\boldsymbol{p}_{g,ij}$ is determined as follows: for each particle $i$, randomly select two other particles from the whole population, compare the fitness of the two particles and choose the better one $\boldsymbol{p}_{g,j}$, $j \ne i$ as the candidate particle. CLPSO [12] determines whether to learn the information of other particles by comparing the random number and the given probability $P_{c,i}$ .

A random number $\varepsilon$ is generated for each dimension and compared with its $P_{c,i}$ . If $\varepsilon$ smaller than $P_{c,i}$ value, the $i$-th particle is guided by other particle's position $\boldsymbol{p}_{g,j}$ . If $\varepsilon$ is larger than $P_{c,i}$ , the particle will follow its own $\boldsymbol{p}_{g,i}$ for that dimension. Therefore, the $\boldsymbol{p}_{g,ij}$ is a new position where each dimension learns from several particles' positions.

We can see from the probability $P_{c,i}$ in [12] that the random numbers are greater than $P_{c,i}$ in most cases. That is, in the updated information of each particle, only a tiny part of the dimensional information comes from other particles' historical best. Interestingly, although the particle in CLPSO updates its position with only a small amount of information from other particles, the search capability is significantly improved.

## 2.3 Niderreiter Theorem

Let $h(\boldsymbol{\theta})$ be a function with the argument $\boldsymbol{\theta} \in E$ ; the modulus of continuity of $h(\boldsymbol{\theta})$ is defined by

$$\omega_h(\varepsilon) = \sup_{\substack{\boldsymbol{\theta}_1, \boldsymbol{\theta}_2 \in E \\ d(\boldsymbol{\theta}_1, \boldsymbol{\theta}_2) \le \varepsilon}} \left| h(\boldsymbol{\theta}_1) - h(\boldsymbol{\theta}_2) \right| , \tag{6}$$

where sup is the abbreviation of supremum, which represents the minimum upper bound in mathematics.

For a sample set $\boldsymbol{P} = [\boldsymbol{\theta}_1, \boldsymbol{\theta}_2, \cdots, \boldsymbol{\theta}_N]$ with $\boldsymbol{\theta}_i \in E$ and $1 \le i \le N$ , the dispersion of $\boldsymbol{P}$ is defined by

$$d_N(\boldsymbol{P}) = \sup_{\boldsymbol{\theta} \in E} \min_{1 \le n \le N} d(\boldsymbol{\theta}, \boldsymbol{\theta}_n) \tag{7}$$

where $d(\cdot, \cdot)$ denotes the euclidean distance. The dispersion is a metric used to measure the uniformity of the sample set. The smaller the dispersion is, the better the uniformity. According to Niderreiter's theorem in [47], we have

$$0 \le \min_{1 \le i \le N} \left( h(\boldsymbol{\theta}_i) \right) - \min_{\boldsymbol{\theta} \in E} \left( h(\boldsymbol{\theta}) \right) \le \omega_h\left( d_N(\boldsymbol{P}) \right) \tag{8}$$

where $\min_{1 \le i \le N} \left( h(\boldsymbol{\theta}_i) \right)$ is the approximate value of best solution in space $E$ , which corresponds to the fitness of the best solution in the field of evolutionary algorithm. $\min_{\boldsymbol{\theta} \in E} \left( h(\boldsymbol{\theta}) \right)$ corresponds to the true optimal solution. Therefore, $\min_{1 \le i \le N} \left( h(\boldsymbol{\theta}_i) \right) - \min_{\boldsymbol{\theta} \in E} \left( h(\boldsymbol{\theta}) \right)$ represents the relative error between the fitness of the best solution and the true optimal solution.

# 3. Effect of population initializers on HCLPSO

In order to explore the relationship between the error of the entire iteration process of PSO and the sample uniformity, this section presents the error analysis on PSO. The error of the result obtained by using PSO is defined by

$$\Delta_g = \min_{1\le i\le N} f\left(\boldsymbol{x}_{g,i}\right) - M \ , \tag{9}$$

where $M$ denotes the exact global optimal fitness value, and $\min_{1\le i\le N} f\left(\boldsymbol{x}_{g,i}\right)$ denotes the optima of all the particle fitness values at the $g$-th iteration. In the error analysis, we assume that PSO can converge to $M$ with iteration.

## 3.1 Error at the initialize iteration

Suppose the position and velocity of the $i$-th particle at the first iteration are

$$\boldsymbol{x}_{0,i} = \boldsymbol{a} + \boldsymbol{\varepsilon}_{1,i} \circ \left(\boldsymbol{b} - \boldsymbol{a}\right), \quad \boldsymbol{v}_{0,i} = \boldsymbol{v}_{\min} + \boldsymbol{\varepsilon}_{2,i} \circ \left(\boldsymbol{v}_{\max} - \boldsymbol{v}_{\min}\right), \tag{10}$$

where

$$\begin{aligned} &\boldsymbol{a} = \left(a_1, a_2, \cdots, a_D\right)^{\mathrm{T}}, \quad \boldsymbol{b} = \left(b_1, b_2, \cdots, b_D\right)^{\mathrm{T}} \\ &\boldsymbol{v}_{\max} = \left(v_{\max,1}, v_{\max,2}, \cdots, v_{\max,D}\right)^{\mathrm{T}} \qquad , \\ &\boldsymbol{v}_{\min} = \left(v_{\min,1}, v_{\min,2}, \cdots, v_{\min,D}\right)^{\mathrm{T}} \end{aligned} \tag{11}$$

$v_{\max,i}$ and $v_{\min,i}$ are the maximum and minimum velocities, respectively, and $\boldsymbol{\varepsilon}_{1,i}$ and $\boldsymbol{\varepsilon}_{2,i}$ are two random vectors in $E := [0,1]^D$ .

Considering (10), $f\left(\boldsymbol{x}_{0,i}\right) = f\left(\boldsymbol{a} + \boldsymbol{\varepsilon}_{1,i} \circ \left(\boldsymbol{b} - \boldsymbol{a}\right)\right)$ , which means the fitness value $f\left(\boldsymbol{x}_{0,i}\right)$ is actually a function of $\boldsymbol{\varepsilon}_{1,i}$ , denoted by $h_0\left(\boldsymbol{\varepsilon}_{1,i}\right) := f\left(\boldsymbol{a} + \boldsymbol{\varepsilon}_{1,i} \circ \left(\boldsymbol{b} - \boldsymbol{a}\right)\right)$. The $\boldsymbol{\varepsilon}_{1,i}$ of all the particles form a sample set $\boldsymbol{P}_0 = \left[\boldsymbol{\varepsilon}_{1,1}, \boldsymbol{\varepsilon}_{1,2}, \cdots, \boldsymbol{\varepsilon}_{1,N}\right]$. According to Niderreiter's theorem(8), the error at the first iteration satisfies

$$\min_{1\le i\le N} h_0\left(\boldsymbol{\varepsilon}_{1,i}\right) - \min_{\boldsymbol{\varepsilon}\in E_1} h_0\left(\boldsymbol{\varepsilon}\right) \le \omega_{h_0}\left(d_N\left(\boldsymbol{P}_0\right)\right) , \tag{12}$$

$$d_N\left(\boldsymbol{P}_0\right) = \sup_{\boldsymbol{\varepsilon}\in E} \min_{1\le n\le N} d\left(\boldsymbol{\varepsilon}, \boldsymbol{\varepsilon}_{1,n}\right) , \tag{13}$$

where $d_N\left(\boldsymbol{P}_0\right)$ is the dispersion of $\boldsymbol{P}_0$. As $\min_{\boldsymbol{\varepsilon}\in E_1} h_0\left(\boldsymbol{\varepsilon}\right) = M$ , we have

$$\Delta_0 = \min_{1\le i\le N} h_0\left(\boldsymbol{\varepsilon}_{1,i}\right) - \min_{\boldsymbol{\varepsilon}\in E_1} h_0\left(\boldsymbol{\varepsilon}\right) \le \omega_{h_0}\left(d_N\left(\boldsymbol{P}_0\right)\right) \tag{14}$$

Obviously, the better the uniformity of $\boldsymbol{P}_0$, the smaller the dispersion and the smaller the error. According to (14), if a more uniformly distributed sample set is used to generate the initial population, the error at the initialize iteration will be smaller.

## 3.2 Error at the first iteration

According to (3), we have

$$\begin{cases} \boldsymbol{v}_{1,i} = \omega\left[\boldsymbol{v}_{\min} + \boldsymbol{\varepsilon}_{2,i}\left(\boldsymbol{v}_{\max} - \boldsymbol{v}_{\min}\right)\right] \\ +c_2\boldsymbol{\varepsilon}_{0,i}^{(2)} \circ \left[\boldsymbol{p}^{(0)} - \left(\boldsymbol{a} + \boldsymbol{\varepsilon}_{1,i}\left(\boldsymbol{b} - \boldsymbol{a}\right)\right)\right]. \\ \boldsymbol{x}_{1,i} = \boldsymbol{x}_{0,i} + \boldsymbol{v}_{1,i} \end{cases} \tag{15}$$

As shown in (15), for all the particles, the parameters $\boldsymbol{a}, \boldsymbol{b}, \boldsymbol{v}_{\min}, \boldsymbol{v}_{\max}, \omega, c_1, c_2, \boldsymbol{p}^{(0)}$ are the same, and the random vectors $\boldsymbol{\varepsilon}_{1,i}$, $\boldsymbol{\varepsilon}_{2,i}$ and $\boldsymbol{\varepsilon}_{1,i}^{(1)}$ are different. For convenience of discussion, let $\boldsymbol{\lambda}_1 := \{\boldsymbol{a}; \boldsymbol{b}; \boldsymbol{v}_{\min}; \boldsymbol{v}_{\max}; \omega; c_1; c_2; \boldsymbol{p}^{(0)}\}$, and $\boldsymbol{\theta}_i^{(1)} := \left(\boldsymbol{\varepsilon}_{1,i}; \boldsymbol{\varepsilon}_{2,i}; \boldsymbol{\varepsilon}_{0,i}^{(2)}\right)$. Therefore, the fitness value $f\left(\boldsymbol{x}_{1,i}\right)$ can be treated as a function of the random argument $\boldsymbol{\theta}_i^{(1)}$, denoted by $h_1\left(\boldsymbol{\theta}_i^{(1)}\right)$. For each particle $i$, there is a random argument $\boldsymbol{\theta}_i^{(1)}$, and all these random arguments form a sample set $\boldsymbol{P}_1 := [\boldsymbol{\theta}_1^{(1)}, \boldsymbol{\theta}_2^{(1)}, \cdots, \boldsymbol{\theta}_N^{(1)}]$. Similarly, using Niderreiter's theorem yields the following error bound

$$\Delta_1 \le \omega_{h_1}\left(d_N\left(\boldsymbol{P}_1\right)\right), \quad d_N\left(\boldsymbol{P}_1\right) = \sup_{\boldsymbol{\theta}\in[0,1]^{3D}} \min_{1\le i\le N} d\left(\boldsymbol{\theta}, \boldsymbol{\theta}_i^{(1)}\right) \tag{16}$$

where $d_N\left(\boldsymbol{P}_1\right)$ denotes the dispersion of $\boldsymbol{P}_1$. (16) gives the error bound at the second iteration. The better the uniformity of $\boldsymbol{P}_1$ is, the smaller $d_N\left(\boldsymbol{P}_1\right)$ and the smaller the error. $\boldsymbol{P}_1$ contains three groups of sequences, in which $\{\boldsymbol{\varepsilon}_{1,i}\}$ and $\{\boldsymbol{\varepsilon}_{2,i}\}$ are the samples for the positions and velocities of the initial population, and $\left\{\boldsymbol{\varepsilon}_{0,i}^{(2)}\right\}$ is required in the velocity updating at the second iteration. In previous studies, the initial population was often generated by using LDS, and hence, $\{\boldsymbol{\varepsilon}_{1,i}\}$ and $\{\boldsymbol{\varepsilon}_{2,i}\}$ are of low-discrepancy, while $\left\{\boldsymbol{\varepsilon}_{0,i}^{(2)}\right\}$ was still generated randomly, which reduces the uniformity of $\boldsymbol{P}_1$. If $\{\boldsymbol{\varepsilon}_{1,i}\}$, $\{\boldsymbol{\varepsilon}_{2,i}\}$ and $\left\{\boldsymbol{\varepsilon}_{0,i}^{(2)}\right\}$ are all generated by using LDS, the dispersion $d_N\left(\boldsymbol{P}_1\right)$ and the error can be reduced.

## 3.3 Error at the g-th iteration

According to (3), we have

$$\begin{cases} \boldsymbol{v}_{2,i} = \omega \boldsymbol{v}_{1,i} + c_1 \boldsymbol{\varepsilon}_{1,i}^{(1)} \circ \left( \boldsymbol{p}_{1,i}^{(l)} - \boldsymbol{x}_{1,i} \right) + c_2 \boldsymbol{\varepsilon}_{1,i}^{(2)} \circ \left( \boldsymbol{p}^{(1)} - \boldsymbol{x}_{1,i} \right) \\ \boldsymbol{x}_{2,i} = \boldsymbol{x}_{1,i} + \boldsymbol{v}_{2,i} \end{cases}, \tag{17}$$

which shows that the particle position $\boldsymbol{x}_{2,i}$ depends on $\omega$, $c_1$, $c_2$, $\boldsymbol{p}_{1,i}^{(l)}$, $\boldsymbol{p}^{(1)}$, $\boldsymbol{\varepsilon}_{1,i}^{(1)}$, $\boldsymbol{\varepsilon}_{1,i}^{(2)}$, $\boldsymbol{x}_{1,i}$ and $\boldsymbol{v}_{1,i}$. According to the updating formula (15), $\boldsymbol{p}_{1,i}^{(l)}$, $\boldsymbol{x}_{1,i}$ and $\boldsymbol{v}_{1,i}$ depend on $\boldsymbol{a}$, $\boldsymbol{b}$, $\boldsymbol{v}_{\min}$, $\boldsymbol{v}_{\max}$, $\omega$, $c_1$, $c_2$, $\boldsymbol{p}^{(0)}$, $\boldsymbol{\varepsilon}_{1,i}$, $\boldsymbol{\varepsilon}_{2,i}$ and $\left\{\boldsymbol{\varepsilon}_{0,i}^{(2)}\right\}$. For all particles, $\boldsymbol{a}$, $\boldsymbol{b}$, $\boldsymbol{v}_{\min}$, $\boldsymbol{v}_{\max}$, $\omega$, $c_1$, $c_2$, $\boldsymbol{p}^{(0)}$, and $\boldsymbol{p}^{(1)}$ are the same, and $\boldsymbol{\varepsilon}_{1,i}$, $\boldsymbol{\varepsilon}_{2,i}$, $\left\{\boldsymbol{\varepsilon}_{0,i}^{(2)}\right\}$, $\boldsymbol{\varepsilon}_{1,i}^{(1)}$ and $\boldsymbol{\varepsilon}_{1,i}^{(2)}$ are different. Let $\boldsymbol{\lambda}_2 := \left\{\boldsymbol{a}; \boldsymbol{b}; \boldsymbol{v}_{\min}; \boldsymbol{v}_{\max}; \omega; c_1; c_2; \boldsymbol{p}^{(0)}; \boldsymbol{p}^{(1)}\right\}$, and $\boldsymbol{\theta}_i^{(2)} := \left(\boldsymbol{\varepsilon}_{1,i}; \boldsymbol{\varepsilon}_{2,i}; \boldsymbol{\varepsilon}_{0,i}^{(2)}; \boldsymbol{\varepsilon}_{1,i}^{(1)}; \boldsymbol{\varepsilon}_{1,i}^{(2)}\right)$. Therefore the fitness value $f\left(\boldsymbol{x}_{2,i}\right)$ is a function of the argument $\boldsymbol{\theta}_i^{(2)}$, denoted by $h_2\left(\boldsymbol{\theta}_i^{(2)}\right)$. Similarly, using Niderreiter's theorem yields the error bound at the third iteration

$$\Delta_2 \le \omega_{h_2}\left(d_N\left(\boldsymbol{P}_2\right)\right), \tag{18}$$

$\boldsymbol{P}_2 = \left[\boldsymbol{\theta}_1^{(2)}, \boldsymbol{\theta}_2^{(2)}, \cdots, \boldsymbol{\theta}_N^{(2)}\right]$ and $d_N\left(\boldsymbol{P}_2\right) = \sup_{\boldsymbol{\theta} \in [0,1]^{5D}} \min_{1 \le i \le N} d\left(\boldsymbol{\theta}, \boldsymbol{\theta}_i^{(2)}\right)$ is the dispersion of $\boldsymbol{P}_2$. The smaller $d_N\left(\boldsymbol{P}_2\right)$ is, the smaller $\Delta_2$ is. Using the same analysis process, the error bound at the $g$-th iteration can be expressed as

$$\begin{aligned} &\Delta_g \le \omega_{h_g}\left(d_N\left(\boldsymbol{P}_g\right)\right) \\ &d_N\left(\boldsymbol{P}_g\right) = \sup_{\boldsymbol{\theta} \in [0,1]^{(2g+1)D}} \min_{1 \le i \le N} d\left(\boldsymbol{\theta}, \boldsymbol{\theta}_i^{(g)}\right) \end{aligned}, \tag{19}$$

where

$$\begin{aligned} &\boldsymbol{P}_g = \left[\boldsymbol{\theta}_1^{(g)}, \boldsymbol{\theta}_2^{(g)}, \cdots, \boldsymbol{\theta}_N^{(g)}\right] \\ &\boldsymbol{\theta}_i^{(g)} = \left(\boldsymbol{\varepsilon}_{1,i}; \boldsymbol{\varepsilon}_{2,i}; \boldsymbol{\varepsilon}_{0,i}^{(2)}; \boldsymbol{\varepsilon}_{1,i}^{(1)}; \boldsymbol{\varepsilon}_{1,i}^{(2)}; \boldsymbol{\varepsilon}_{2,i}^{(1)}; \boldsymbol{\varepsilon}_{2,i}^{(2)}; \cdots; \boldsymbol{\varepsilon}_{g-1,i}^{(1)}; \boldsymbol{\varepsilon}_{g-1,i}^{(2)}\right) \end{aligned}. \tag{20}$$

(19) shows that the error bound $\Delta_g$ depends on the dispersion of $\boldsymbol{P}_g$, which is a sample set in the $(2g+1)D$ -dimensional space.

# 4. PSO With Low-discrepancy Sampling in the Expanded Dimensional Space

In this section, a new acceleration technique, LDS in the expanded dimensional space (LDSEDS) is given. According to the different sample construction methods, two versions of LDSEDS are proposed. These two techniques are combined with PSO to generate two more effective algorithms.

## 4.1 Inspiration from Error Analysis

In the introduction, we raised two questions. The first being, why can't the advantages of using LDS initialization at the early iterations be preserved during the entire iterations of the PSO? Furthermore, what is the relationship between the error of the entire iteration process of PSO and the sample uniformity? The error analysis in Section 3 explains the second question. As shown in (19), the error bound of PSO at each iteration is dependent on the dispersion of the sample set $\boldsymbol{P}_g$. The answer to the first question is also revealed; in the iterative process of PSO, it is necessary to sample dynamically in the space with increasing dimensions. Generating the initial population with LDS only ensures that the sample sets $\{\varepsilon_{1,i}\}$ and $\{\varepsilon_{2,i}\}$ in the first two dimensions of $\boldsymbol{P}_g$ are of low-discrepancy but ignore the other dimensions. With the increase in iterations, $\boldsymbol{P}_g$ contains an increasing number of random numbers, becoming a random sample set, and the advantages introduced by the low-discrepancy of $\{\varepsilon_{1,i}\}$ and $\{\varepsilon_{2,i}\}$ are completely destroyed.

At the same time, the error analysis in Section 3 also explains why the effect of using LDS in each iteration step is worse. This is because the existing LDSs are mostly deterministic point sets, so the uniformity of the whole iteration process is destroyed, so the calculation error of the algorithm increases, resulting in the worse calculation results.

PSO requires the sample set $\boldsymbol{P}_g$ in $(2g+1)D$ -dimensional space. Suppose we use LDS instead of random samples to generate $\boldsymbol{P}_g$. In that case, the dispersion of $\boldsymbol{P}_g$ will be reduced, the error of PSO at each iteration will be reduced and the convergence speed will be improved. This is the critical technique of the accelerated PSO proposed in this paper, which we name low-discrepancy sampling in the expanded dimensional space (LDSEDS). The PSO acceleration algorithm combined with this technique is called PSO with LDS in the expanded dimensional space (PSO-LDSEDS).

## 4.2 Low-discrepancy Sampling in Expanded Dimensional Space Based on Number Theory

The LDSEDS based on number theory directly generates the low-discrepancy sample set $\boldsymbol{P}_G$ in the ultrahigh dimension. The dimension of $\boldsymbol{P}_G$ is so high that some LDSs based on optimization algorithms, e.g., DES and OHS, are unsuitable because of the enormous computational cost required to generate the samples. Some LDSs based on number theory stand out in this case, such as HWS, SS and HSS. For such high-dimensional space, we used HSS, a modified version of the famous HS. By adding some perturbation to the original HS, HSS alleviates the strip distribution of the sample set form which HS, as well as HWS and SS, suffers.

Then, this technique is combined with PSO. If the maximum number of iterations is $G$, the iteration requires $N$ samples $\boldsymbol{P}_G := \left[\boldsymbol{\theta}_1^{(G)}, \cdots, \boldsymbol{\theta}_N^{(G)}\right]$ in $(2G+1)D$ dimensional space. The first $2D$ dimensional numbers in $\boldsymbol{P}_G$ are used for the generation of the initial positions and velocities. The numbers in dimensions $2g$ and $2g+1$ are used to generate $\boldsymbol{\varepsilon}_g^{(1)}$ and $\boldsymbol{\varepsilon}_g^{(2)}$, respectively. This improved PSO algorithm is called PSO-LDSEDS based on number theory (PSO-LDSEDS1).

## 4.3 Low-discrepancy Sampling in Expanded Dimensional Space Based on a Combination of Low-dimensional Samples

Generally, the difference between the low-discrepancy sample set generated by number theory and that of a random sample set increases with the number of samples but decreases with the dimension. Considering that the dimension of the sample set $\boldsymbol{P}_G$ involved in PSO is so high that directly generating it with number theory method may not decrease its discrepancy when the number of particles is not too large. Therefore, we propose another LDSEDS based on a combination of low-dimensional samples.

Supposing we have obtained a low-discrepancy sample set $\boldsymbol{\Theta}$ with $N$ samples in $D$-

dimensional space,

$$\boldsymbol{\Theta}=\begin{bmatrix}\theta_{11} & \theta_{12} & \cdots & \theta_{1N}\\ \theta_{21} & \theta_{22} & \cdots & \theta_{2N}\\ \vdots & \vdots & \ddots & \vdots\\ \theta_{D1} & \theta_{D2} & \cdots & \theta_{DN}\end{bmatrix}=\begin{bmatrix}\boldsymbol{\theta}_{1,:}\\ \boldsymbol{\theta}_{2,:}\\ \vdots\\ \boldsymbol{\theta}_{D,:}\end{bmatrix}, \tag{21}$$

and now we want to generate $N$ samples in $2D$-dimensional space. For the number array $[1, 2, \cdots, D]$, we make a random permutation, denoted by $\boldsymbol{\pi}=[\pi_1, \pi_2, \cdots, \pi_D]$. Using $\boldsymbol{\pi}$ can yield a new low-discrepancy sample set $\boldsymbol{\Theta}_{\pi}$

$$\boldsymbol{\Theta}_{\pi}=\begin{bmatrix}\theta_{\pi_1 1} & \theta_{\pi_1 2} & \cdots & \theta_{\pi_1 N}\\ \theta_{\pi_2 1} & \theta_{\pi_2 2} & \cdots & \theta_{\pi_2 N}\\ \vdots & \vdots & \ddots & \vdots\\ \theta_{\pi_D 1} & \theta_{\pi_D 2} & \cdots & \theta_{\pi_D N}\end{bmatrix}=\begin{bmatrix}\boldsymbol{\theta}_{\pi_1,:}\\ \boldsymbol{\theta}_{\pi_2,:}\\ \vdots\\ \boldsymbol{\theta}_{\pi_D,:}\end{bmatrix}. \tag{22}$$

Clearly, $[\boldsymbol{\Theta};\boldsymbol{\Theta}_{\boldsymbol{\pi}}]$ forms a new low-discrepancy sample set with $N$ samples in $2D$-dimensional space. Then, we apply this technique to PSO. Using this combining method can easily yield $\boldsymbol{P}_G$ in $(2G+2)\times D$-dimensional space,

$$\boldsymbol{P}_G=[\boldsymbol{\Theta}; \boldsymbol{\Theta}_{\boldsymbol{\pi}_1}; \cdots; \boldsymbol{\Theta}_{\boldsymbol{\pi}_{2G+2}}], \tag{23}$$

where $\boldsymbol{\pi}_i$ represents the random permutation. By combining the low-dimensional samples generated using different LDSs (such as HWS, DES, HSS and SS) in a random permutational way, it is quite convenient to generate the samples in the dynamically expanded dimensional space. This more efficient PSO algorithm is called PSO-LDSEDS, and is based on a combination of low-dimensional samples (PSO-LDSEDS2).

# 5. Numerical Test of PSO-LDSEDS

This section uses numerical experiments to verify the correctness of the error analysis in Section 3 and the effectiveness of the two improved algorithms proposed in Section 4.

## 5.1 Measurement of Performance

In this section, the PSOs with two different samplings are considered as two different algorithms. To compare the performance of the different algorithms, we used the following metrics:

Error tolerance ($\varepsilon_{\text{tol}}$): When the relative error between the fitness of the best solution and the true optimal solution is smaller than $\varepsilon_{\text{tol}}$, the algorithm is considered convergent.

Convergence speed (CS): This metric indicates the number of iterations after which the algorithm converges. For the two algorithms, the smaller CS is, the more powerful the search ability of the algorithm. Due to the randomness existing in PSO, 60 runs are performed for each test function and for each algorithm. The average convergence curve of 60 runs is recorded, and the CS for a given error tolerance is evaluated based on the average convergence curve. If the CS is greater than the maximum number of iterations $G$, it is claimed that the algorithm fails to find the global optimum under the given error tolerance, and it is represented by "-" in the table.

## 5.2 Test Functions

In this section, 15 benchmark functions selected from the 30 widely used benchmark functions provided by the CEC 2017 special session [48] are used as the test functions. The name, topology, and true optimal solutions ($Z^*$) of these 15 test functions are listed in Table I. For the other 15 functions the original PSO cannot converge within $G = 7500$ iterations, and they are discarded.

TABLE I

BENCHMARK FUNCTIONS

| Topology | No. | Function name | $Z^*$ |
|---|---|---|---|
| Unimodal Fns. | $F_1$ | Shifted and rotated Zakharov function | 300 |
| Simple multimodal Fns. | $F_2$ | Shifted and rotated Rosenbrock's function | 400 |
| | $F_3$ | Shifted and rotated Rastrigin's function | 500 |
| | $F_4$ | Shifted and rotated Expaned Schaffer F6 function | 600 |
| | $F_5$ | Shifted and rotated Lunacek Bi-Ratrigin's function | 700 |
| | $F_6$ | Shifted and rotated Non-continuous Rastrigin's function | 800 |
| | $F_7$ | Shifted and rotated Lvey function | 900 |
| Hybrid Fns. | $F_8$ | Zakharov, Rosenbrock, Rastrigin | 1100 |
| | $F_9$ | High-conditioned elliptic; Ackley; Schaffer; Rastrigin | 1400 |
| | $F_{10}$ | Bent Cigar; HGBat; Rastrigin; Rosenbrock | 1500 |
| | $F_{11}$ | Expanded schaffer; HGBat; Rosenbrock; Modified Schwefel; Rastrigin | 1600 |
| | $F_{12}$ | Katsuura; Ackely; Expanded Griewank plus Rosenbrock; Schwefel; Rastrigin | 1700 |
| | $F_{13}$ | High-conditioned elliptic; Ackley; Rastrigin; HGBat; Dicus | 1900 |
| | $F_{14}$ | Bent Cigar; Griewank plus Rosenbrock; Rastrigin; Expanded Schaffer | 2000 |
| Composite Fns. | $F_{15}$ | Griewank; Rastrigin; Modified schwefel | 2200 |

Fns. Functions

## 5.3 Statistical Analysis

The modified Friedman test (MFT) [49] and the Nemenyi test (NT) [50] were used here to compare the performance of the different algorithms. The MFT justifies whether there are significant differences between the considered algorithms in terms of the Friedman statictics $\tau_F$

$$\tau_F = \frac{(m-1)\chi_F^2}{m(k-1)-\chi_F^2}, \quad \chi_F^2 = \frac{12m}{k(k+1)}\left[\sum_i R_i^2 - \frac{k(k+1)^2}{4}\right], \tag{24}$$

where $m$ is the number of test functions, $k$ is the number of algorithms considered, and $R_i$ is the average rank of the $i$-th algorithm. The smaller $R_i$ is, the better the performance of the $i$-th algorithm. If $\tau_F$ is greater than the critical value $\tau_c$ for a given significance level $\alpha$, it is claimed that there are significant differences between the considered algorithms. If significant differences are observed, the NT is used to justify which algorithm is significantly superior to the other algorithms. This test states that the performance of the two algorithms is significantly different if the corresponding average ranks differ by at least a critical difference, defined as follows:

$$CD = q_\alpha \sqrt{\frac{k(k+1)}{6m}}, \tag{25}$$

where $q_\alpha$ can be obtained by querying the studentized range statistic table [50] with the significance level $\alpha$. In all the experiments, the confidence level is set to $\alpha = 0.05$.

## 5.4 Performance of PSO-LDSEDS1

In this section, a comparison between PSO and PSO-LDSEDS1 is presented using the test functions with $D=10$. The parameters involved are the same for both algorithms except for the sample sets. The maximum number of iterations is $G = 7500$. $w$ decreases linearly with iterations in the range of 0.9-0.4. $c_1$ and $c_2$ vary linearly with iterations in the range of 2.5-0.5 and 0.5-2.5, respectively. To compare the CSs of different algorithms at different error tolerances, $\varepsilon_{\text{tol}} = 1\%$ and $5\%$ are used.

PSO-LDSEDS1 directly uses the HSS method to generate $\boldsymbol{P}_G$ with dimensions of $(2G+2)D$. Obviously, $\boldsymbol{P}_G$ is an ultrahigh dimensional LDS, so it may need a large number of samples to match the dimensions to achieve the ideal uniformity effect. Here, three groups of comparative experiments with $N = 40$, 100, and 160, are given to compare the CSs of the original PSO (denoted by Rand) and PSO-LDSEDS1 (denoted by HSS) at $\varepsilon_{\text{tol}} = 1\%$ and $5\%$. The CSs and ranks of Rand and HSS with different population sizes at different tolerance errors and population sizes are listed in Table II.

As shown in Table II, the AvgRks of PSO-LDSEDS1 are smaller than those of PSO in almost all simulation conditions, except $N = 40$ and $\varepsilon_{\text{tol}} = 5\%$. For both $\varepsilon_{\text{tol}} = 1\%$ and $5\%$, the Friedman statistics $\tau_F$ increase as $N$ increases. This is because the LDS covers the search space better than the random sample set, and the advantage becomes more evident as the population size increases. When $N = 160$, the Friedman statistics at $\varepsilon_{\text{tol}} = 1\%$ and $5\%$ are both far greater than the critical value $\tau_c = 4.600$. Therefore, it can be stated that when $N = 160$, PSO-LDSEDS1 converges significantly faster than PSO at the significance level $\alpha = 0.05$.

The experimental result shows that, although directly generating the low-discrepancy sample $\boldsymbol{P}_G$ can improve the numerical performance of PSO, the population size required needs to be large enough, which increases the computational cost. Next, we test the performance of PSO-LDSEDS2.

TABLE II

CSS AND RANKS OF DIFFERENT ALGORITHMS WITH DIFFERENT POPULATION SIZES AT DIFFERENT TOLERANCE ERRORS

| | $\varepsilon_{\text{tol}} = 5\%$ | | | | | | $\varepsilon_{\text{tol}} = 1\%$ | | | | | |
|---|---|---|---|---|---|---|---|---|---|---|---|---|
| | $N = 40$ | | $N = 100$ | | $N = 160$ | | $N = 40$ | | $N = 100$ | | $N = 160$ | |
| | Rand | HSS | Rand | HSS | Rand | HSS | Rand | HSS | Rand | HSS | Rand | HSS |
| $F_1$ | 1515(2) | 1281(1) | 1385(2) | 1223(1) | 1336(2) | 1201(1) | 1626(2) | 1388(1) | 1492(2) | 1314(1) | 1436(2) | 1289(1) |
| $F_2$ | 189(1) | 353(2) | 82(1) | 117(2) | 52(2) | 42(1) | 2006(2) | 1611(1) | 1799(2) | 1563(1) | 1688(2) | 1406(1) |
| $F_3$ | 948(2) | 747(1) | 835(2) | 811(1) | 745(1) | 770(2) | -(1.5) | -(1.5) | 2765(1) | 4674(2) | 2252(1) | 2306(2) |
| $F_4$ | 24(2) | 17(1) | 16(2) | 15(1) | 15(2) | 13(1) | 474(1) | 479(2) | 314(1) | 346(2) | 254(2) | 233(1) |
| $F_5$ | 1420(2) | 1098(1) | 1299(2) | 1072(1) | 1244(2) | 1048(1) | -(1.5) | -(1.5) | -(1.5) | -(1.5) | -(1.5) | -(1.5) |
| $F_6$ | 582(2) | 494(1) | 327(1) | 454(2) | 325(2) | 287(1) | 2472(1) | -(2) | 1928(1) | 2008(2) | 1716(2) | 1683(1) |
| $F_7$ | 562(2) | 511(1) | 418(2) | 415(1) | 281(2) | 272(1) | 908(2) | 729(1) | 803(2) | 704(1) | 675(2) | 633(1) |
| $F_8$ | 350(1) | 398(2) | 228(2) | 208(1) | 172(2) | 160(1) | 1061(2) | 905(1) | 914(2) | 821(1) | 860(2) | 809(1) |
| $F_9$ | 1054(1) | 1121(2) | 753(2) | 739(1) | 705(2) | 632(1) | -(1.5) | -(1.5) | 5713(1) | -(2) | 2715(2) | 2651(1) |
| $F_{10}$ | 1995(1) | 2070(2) | 1394(2) | 1283(1) | 1305(2) | 1205(1) | 3767(1) | -(2) | 2588(2) | 2559(1) | 2310(2) | 2128(1) |
| $F_{11}$ | 452(1) | 460(2) | 233(1) | 261(2) | 191(2) | 169(1) | 906(2) | 843(1) | 601(1) | 683(2) | 601(2) | 563(1) |
| $F_{12}$ | 392(2) | 328(1) | 162(1) | 208(2) | 137(2) | 134(1) | 1730(2) | 1720(1) | 1360(2) | 1286(1) | 1261(2) | 1243(1) |
| $F_{13}$ | 1656(1) | -(2) | 1019(2) | 888(1) | 793(2) | 774(1) | 3230(1) | -(2) | 1726(2) | 1703(1) | 1552(2) | 1526(1) |
| $F_{14}$ | 165(2) | 147(1) | 83(1) | 90(2) | 65(2) | 64(1) | 1200(2) | 1140(1) | 1116(2) | 972(1) | 956(2) | 893(1) |
| $F_{15}$ | 814(2) | 723(1) | 396(1) | 458(2) | 200(2) | 22(1) | -(1.5) | -(1.5) | -(1.5) | -(1.5) | -(2) | 736(1) |
| AvgR. | 1.60(2) | 1.40(1) | 1.60(2) | 1.40(1) | 1.93(2) | 1.07(1) | 1.60(2) | 1.40(1) | 1.60(2) | 1.40(1) | 1.90(2) | 1.10(1) |
| $\tau_F$ | 0.583 | | 0.583 | | 42.250 | | 0.583 | | 0.583 | | 24.889 | |

AvgR. Average ranks

## 5.5 Performance of PSO-LDSEDS2

The comparison between PSO and PSO-LDSEDS2 using the $D = 10$ dimensional test functions is presented first. The adopted parameters are the same as those used previously, except for the sample sets $N = 40$. PSO-LDSEDS2 uses the combining method introduced in Section 4 to generate $\boldsymbol{P}_G$. As shown by (23), the seed sample set $\boldsymbol{\Theta}$ was generated using four LDSs, i.e., HWS, DES, HSS, and SS. For convenience, PSO-LDSEDS2, which uses certain types of LDSs to generate the seed sample set, is denoted by the name of the corresponding LDS. As the original PSO generates $\boldsymbol{P}_G$ by random sampling, it is denoted by Rand. The CSs, the computation times (CTs) and their ranks of Rand, HWS, DES, HSS, and SS at different tolerance errors are listed in Table III and Table IV.

The comparison in Table III and Table IV shows that under the same accuracy requirements, the number of iterations and the CTs required by PSO-LDSEDS2 is significantly less than that required by PSO. According to Table III and Table IV, for both $\varepsilon_{\text{tol}} = 5\%$ and 1%, $\tau_F$ is greater than the critical value $\tau_{\text{c}} = 2.537$ in the significance level $\alpha = 0.05$. According to NT and (25), the critical difference $CD = 1.575$. Therefore, the AvgRks of PSO-LDSEDS2 with different LDSs are always significantly smaller than those of PSO at both $\varepsilon_{\text{tol}} = 5\%$ and 1%.

Next, the performance of PSO-LDSEDS2 is evaluated using the $D = 30$ dimensional test functions. When $D = 30$, there are many test functions in Table I for which both PSO and PSO-LDSEDS2 fail to find solutions with relative errors smaller than $\varepsilon_{\text{tol}} = 5\%$. Therefore, $\varepsilon_{\text{tol}} = 20\%$ and $\varepsilon_{\text{tol}} = 10\%$ were used. In addition, there are still five test functions for which the adopted algorithms fail at $\varepsilon_{\text{tol}} = 20\%$, and we adopted only the remaining ten test functions. The population size is $N = 100$. The other parameters are the same as those used previously. The CSs, the CTs and their ranks of the considered algorithms at different tolerance errors are listed in Table V and Table VI.

The comparison in Table V and Table VI shows that for 30-dimensional problems, PSO-LDSEDS2 still performs significantly better than PSO under the same error tolerance. For both $\varepsilon_{\text{tol}} = 20\%$ and 10%, $\tau_F$ are greater than the critical value $\tau_{\text{c}} = 2.634$ in the significance level $\alpha = 0.05$. Therefore, there are significant differences between the CSs and the CTs of the five algorithms considered. The AvgRks of HWS, DES, HSS and SS are all smaller than that of Rand.

According to NT and (25), $CD = 1.929$. Therefore, HWS, DES and HSS perform significantly better than Rand at $\varepsilon_{\text{tol}} = 20\%$, and HWS and DES perform significantly better than Rand at $\varepsilon_{\text{tol}} = 10\%$ .Although the difference between the AvgRks of Rand and SS is not greater than $CD$, SS actually performs better than Rand for almost all the test functions.

TABLE III

CSs AND RANKS OF THE FIVE ALGORITHMS AT DIFFERENT TOLERANCE ERRORS

| | $\varepsilon_{\text{tol}} = 5\%$ | | | | | $\varepsilon_{\text{tol}} = 1\%$ | | | | |
|---|---|---|---|---|---|---|---|---|---|---|
| | Rand | HWS | DES | HSS | SS | Rand | HWS | DES | HSS | SS |
| $F_1$ | 1515(5) | 943(2) | 977(4) | 963(3) | 916(1) | 1626(5) | 1032(2) | 1064(4) | 1056(3) | 997(1) |
| $F_2$ | 189(5) | 62(1) | 142(4) | 94(3) | 73(2) | 2006(5) | 1532(2) | 1800(3) | 1813(4) | 1451(1) |
| $F_3$ | 948(5) | 428(1) | 496(3) | 520(4) | 488(2) | -(5) | 4202(4) | 3474(3) | 2347(1) | 3271(2) |
| $F_4$ | 24(5) | 20(2) | 21(3) | 19(1) | 22(4) | 474(5) | 128(1) | 136(3) | 149(4) | 130(2) |
| $F_5$ | 1420(5) | 856(2) | 804(1) | 870(3) | 901(4) | -(3) | -(3) | -(3) | -(3) | -(3) |
| $F_6$ | 582(5) | 192(2) | 195(3) | 179(1) | 197(4) | 2472(5) | 1717(2) | 1786(4) | 1702(1) | 1724(3) |
| $F_7$ | 562(5) | 163(2) | 178(4) | 171(3) | 152(1) | 908(5) | 381(3) | 384(4) | 376(2) | 335(1) |
| $F_8$ | 350(5) | 169(3) | 154(1.5) | 154(1.5) | 196(4) | 1061(5) | 625(3) | 610(1) | 652(4) | 612(2) |
| $F_9$ | 1054(5) | 833(1) | 874(2) | 941(4) | 899(3) | -(3) | -(3) | -(3) | -(3) | -(3) |
| $F_{10}$ | 1995(4) | 2068(5) | 1841(3) | 1766(2) | 1747(1) | 3767(5) | 3666(2) | 3633(1) | 3765(4) | 3668(3) |
| $F_{11}$ | 452(5) | 308(3) | 280(2) | 345(4) | 263(1) | 906(5) | 831(4) | 672(2) | 749(3) | 624(1) |
| $F_{12}$ | 392(5) | 229(3) | 251(4) | 207(2) | 168(1) | 1730(3) | 2328(4) | 2947(5) | 1676(1) | 1702(2) |
| $F_{13}$ | 1656(4) | 1499(1) | 1646(3) | 1501(2) | 2115(5) | 3230(4) | 3185(3) | 3127(2) | 3030(1) | 4001(5) |
| $F_{14}$ | 165(5) | 158(4) | 121(2) | 118(1) | 124(3) | 1200(5) | 976(3) | 858(1) | 914(2) | 979(4) |
| $F_{15}$ | 814(5) | 193(3) | 256(4) | 131(1) | 140(2) | -(3) | -(3) | -(3) | -(3) | -(3) |
| AvgR. | 4.87(5) | 2.33(1) | 2.90(4) | 2.37(2) | 2.53(3) | 4.40(5) | 2.800(3.5) | 2.800(3.5) | 2.600(2) | 2.400(1) |
| $\tau_F$ | | | 11.728 | | | | | 4.817 | | |

TABLE IV

CTs AND RANKS OF THE FIVE ALGORITHMS AT DIFFERENT TOLERANCE ERRORS

| | $\varepsilon_{\text{tol}} = 5\%$ | | | | | $\varepsilon_{\text{tol}} = 1\%$ | | | | |
|---|---|---|---|---|---|---|---|---|---|---|
| | Rand | HWS | DES | HSS | SS | Rand | HWS | DES | HSS | SS |
| $F_1$ | 0.101(5) | 0.064(1) | 0.070(3) | 0.071(4) | 0.065(2) | 0.109(5) | 0.069(2) | 0.071(3) | 0.074(4) | 0.067(1) |
| $F_2$ | 0.010(4) | 0.004(1) | 0.014(5) | 0.006(3) | 0.004(2) | 0.097(5) | 0.069(1) | 0.086(3) | 0.091(4) | 0.074(2) |
| $F_3$ | 0.064(5) | 0.033(1) | 0.034(3) | 0.038(4) | 0.033(2) | 0.321(5) | 0.229(1) | 0.256(3) | 0.247(2) | 0.284(4) |
| $F_4$ | 0.004(5) | 0.003(1) | 0.004(4) | 0.003(2) | 0.004(3) | 0.060(5) | 0.017(2) | 0.019(4) | 0.019(3) | 0.017(1) |
| $F_5$ | 0.101(5) | 0.054(1) | 0.061(3) | 0.067(4) | 0.056(2) | 0.592(5) | 0.570(2) | 0.543(1) | 0.591(4) | 0.586(3) |
| $F_6$ | 0.041(5) | 0.016(4) | 0.016(3) | 0.015(2) | 0.012(1) | 0.234(5) | 0.157(2) | 0.152(1) | 0.162(3) | 0.173(4) |
| $F_7$ | 0.041(5) | 0.010(1) | 0.011(2) | 0.015(4) | 0.011(3) | 0.065(5) | 0.025(2) | 0.026(3) | 0.029(4) | 0.024(1) |
| $F_8$ | 0.022(5) | 0.010(2) | 0.009(1) | 0.011(3) | 0.012(4) | 0.068(5) | 0.034(1) | 0.040(2) | 0.054(4) | 0.044(3) |
| $F_9$ | 0.076(5) | 0.064(2) | 0.059(1) | 0.074(4) | 0.066(3) | 0.443(5) | 0.403(1) | 0.426(3) | 0.428(4) | 0.404(2) |
| $F_{10}$ | 0.148(4) | 0.137(2) | 0.152(5) | 0.136(1) | 0.145(3) | 0.255(5) | 0.214(1) | 0.229(2) | 0.239(4) | 0.230(3) |
| $F_{11}$ | 0.030(4) | 0.023(3) | 0.017(1) | 0.030(5) | 0.020(2) | 0.062(5) | 0.040(1) | 0.042(2) | 0.048(4) | 0.043(3) |
| $F_{12}$ | 0.042(5) | 0.031(3) | 0.036(4) | 0.024(1) | 0.030(2) | 0.642(4) | 0.654(5) | 0.521(3) | 0.442(1) | 0.446(2) |
| $F_{13}$ | 0.514(3) | 0.368(1) | 0.535(4) | 0.489(2) | 0.630(5) | 1.208(3) | 1.114(1) | 1.215(4) | 1.127(2) | 1.281(5) |
| $F_{14}$ | 0.024(5) | 0.022(4) | 0.015(2) | 0.017(3) | 0.015(1) | 0.319(5) | 0.259(2) | 0.206(1) | 0.261(3) | 0.317(4) |
| $F_{15}$ | 0.125(5) | 0.044(3) | 0.051(4) | 0.042(2) | 0.039(1) | 1.293(4) | 1.311(5) | 1.276(3) | 1.212(1) | 1.232(2) |
| AvgR. | 4.667(5) | 2.000(1) | 3.000(4) | 2.933(3) | 2.400(2) | 4.733(5) | 1.933(1) | 2.533(2) | 3.133(4) | 2.667(3) |
| $\tau_F$ | | | 9.900 | | | | | 11.430 | | |

TABLE V

CSs AND RANKS OF THE FIVE ALGORITHMS AT DIFFERENT TOLERANCE ERRORS

| | $\varepsilon_{\text{tol}} = 20\%$ | | | | | $\varepsilon_{\text{tol}} = 10\%$ | | | | |
|---|---|---|---|---|---|---|---|---|---|---|
| | Rand | HWS | DES | HSS | SS | Rand | HWS | DES | HSS | SS |
| $F_1$ | 3702(5) | 3491(3) | 3456(1) | 3487(2) | 3545(4) | 3787(5) | 3595(3) | 3571(1) | 3572(2) | 3644(4) |
| $F_3$ | 1843(5) | 1498(3) | 1455(1) | 1497(2) | 1519(4) | 2803(5) | 2011(2) | 2228(4) | 2048(3) | 1949(1) |
| $F_4$ | 2(3) | 2(3) | 2(3) | 2(3) | 2(3) | 21(5) | 20(3.5) | 19(1.5) | 19(1.5) | 20(3.5) |
| $F_5$ | 1985(5) | 1581(1) | 1598(3) | 1596(2) | 1704(4) | -(3) | -(3) | -(3) | -(3) | -(3) |
| $F_6$ | 1665(5) | 1122(2) | 1195(4) | 1120(1) | 1139(3) | 2016(5) | 1580(2) | 1606(3) | 1538(1) | 1647(4) |
| $F_7$ | 1691(5) | 1371(4) | 1357(3) | 1289(1) | 1356(2) | 2020(5) | 1744(2) | 1852(3) | 1725(1) | 1963(4) |
| $F_8$ | 1409(5) | 1105(3) | 1066(1) | 1122(4) | 1079(2) | 2001(2) | 1963(1) | 2595(3) | 2602(4) | 2703(5) |
| $F_{12}$ | 1168(5) | 970(2) | 956(1) | 988(3) | 1033(4) | 2962(5) | 2275(3) | 2000(2) | 1856(1) | 2633(4) |
| $F_{14}$ | 1038(5) | 732(1) | 770(2) | 1032(4) | 857(3) | 1761(4) | 1443(2) | 1394(1) | 3224(5) | 1636(3) |
| $F_{15}$ | 823(5) | 324(2) | 188(1) | 387(4) | 375(3) | 1402(5) | 531(2) | 310(1) | 940(4) | 553(3) |
| AvgR. | 4.800(5) | 2.400(2) | 2.000(1) | 2.600(3) | 3.200(4) | 4.400(5) | 2.350(2) | 2.250(1) | 2.550(4) | 3.450(3) |
| $\tau_F$ | | | 8.308 | | | | | 4.534 | | |

TABLE VI

CTs AND RANKS OF THE FIVE ALGORITHMS AT DIFFERENT TOLERANCE ERRORS

| | $\varepsilon_{\text{tol}} = 20\%$ | | | | | $\varepsilon_{\text{tol}} = 10\%$ | | | | |
|---|---|---|---|---|---|---|---|---|---|---|
| | Rand | HWS | DES | HSS | SS | Rand | HWS | DES | HSS | SS |
| $F_1$ | 1.866(5) | 1.492(1) | 1.559(2) | 1.566(3) | 1.570(4) | 1.744(5) | 1.423(1) | 1.484(3) | 1.462(2) | 1.487(4) |
| $F_3$ | 1.253(5) | 0.916(2) | 0.792(1) | 0.939(3) | 0.944(4) | 3.095(5) | 2.304(2) | 2.081(1) | 2.465(4) | 2.375(3) |
| $F_4$ | 0.007(2) | 0.007(3) | 0.007(1) | 0.007(4) | 0.008(5) | 0.028(5) | 0.023(2) | 0.022(1) | 0.025(3) | 0.026(4) |
| $F_5$ | 1.648(5) | 1.140(1) | 1.215(2) | 1.222(3) | 1.340(4) | 5.422(5) | 5.055(2) | 5.015(1) | 5.197(4) | 5.078(3) |
| $F_6$ | 1.171(5) | 0.806(3) | 0.624(1) | 0.831(4) | 0.771(2) | 1.636(5) | 1.321(3) | 1.107(1) | 1.151(2) | 1.355(4) |
| $F_7$ | 1.182(5) | 0.873(2) | 0.963(3) | 0.968(4) | 0.829(1) | 1.462(3) | 1.254(1) | 1.742(5) | 1.314(2) | 1.488(4) |
| $F_8$ | 0.913(5) | 0.655(3) | 0.576(1) | 0.761(4) | 0.627(2) | 1.785(2) | 1.827(3) | 1.706(1) | 2.367(5) | 2.127(4) |
| $F_{12}$ | 1.912(5) | 1.338(2) | 1.582(3) | 1.756(4) | 1.273(1) | 4.454(5) | 3.842(1) | 4.097(3) | 3.907(2) | 4.369(4) |
| $F_{14}$ | 1.397(5) | 0.998(1) | 1.031(2) | 1.352(4) | 1.169(3) | 3.323(1) | 3.431(2) | 3.448(3) | 6.032(5) | 3.774(4) |
| $F_{15}$ | 1.952(5) | 0.761(3) | 0.217(1) | 0.834(4) | 0.456(2) | 1.928(5) | 0.773(2) | 0.451(1) | 0.996(4) | 0.838(3) |
| AvgR. | 4.700(5) | 2.100(2) | 1.700(1) | 3.700(4) | 2.800(3) | 4.100(5) | 1.900(1) | 2.000(2) | 3.300(3) | 3.700(4) |
| $\tau_F$ | | | 13.059 | | | | | 6.000 | | |

# 5. Disscussions

From previous experiments, we can conclude that generating $\boldsymbol{P}_G$ with LDSs improves the convergence speed of PSO. For PSO-LDSEDS1, when $N = 40$, its performance is similar to that of PSO. This is because the population size is so small that the uniformity of the high-dimensional sample set generated by directly using the LDS is not better than that of the random sample sets. However, with an increase in the population size, PSO-LDSEDS1 performs increasingly better. When $N = 160$, the convergence speed of PSO-LDSEDS1 is significantly superior to that of PSO. The PSO-LDSEDS2 can substantially improve the convergence speed without reducing the search ability of the original PSO for the 10- and 30-dimensional problems. We adopted four sampling methods to generate the seed low-discrepancy sample set, and HWS and DES are more robust than the other two samplings. Tables III-VI show that the combined sampling technique is more efficient than direct sampling based on number theory, and both techniques are more efficient than random sampling.

It is natural to question whether the LDSEDS technique can be extended to other modified versions of PSO. In the next section, the LDSEDS was combined with a widely used modified PSO, i.e., CLPSO, and experiments and discussions were presented for testing the performance of CLPSO-LDSEDS.

# 6. Extended to CLPSO

It seems that LDSEDS can be extended to other PSO variants. According to the error analysis in Section 3, there is a precondition that the position updating of each particle depends on the random variables that are independent of the other particles and a global best that is the same for all the particles. Nevertheless, among the many variants of PSO, few algorithms can strictly meet these preconditions except for PSO, which directly changes the control parameter. For some PSO variants that change the population topologies, each particle learns a lot from the adjacent particles, which undoubtedly does not meet the preconditions of the analysis in Section 3. However, there are some excellent PSO variants, such as CLPSO and HCLPSO, in which each particle updates its position with only a small amount of information from the other particles and mainly depends on the information of the particle itself. These types of algorithms could be expected to accelerate the convergence speed by using the LDSEDS. In this section, we choose the classic CLPSO among these variants as the carrier for the next promotion.

As shown in Section 5, the combining sampling based on random permutations of a low-dimensional low-discrepancy sample set is more efficient, and it will be adopted for CLPSO.

Unlike PSO, the velocity updating formula of CLPSO does not involve global optima. Therefore, in each iteration of CLPSO, it is only necessary to sample the random vector $\boldsymbol{\varepsilon}_{g,i}$ once. Within $G$ iterations, the dimension of the total sample set $\boldsymbol{P}_G$ is $(G+1)\times D$. For a seed low-discrepancy sample set $\boldsymbol{\Theta}$, combining its random permutations yields

$$\boldsymbol{P}_G = [\boldsymbol{\Theta}; \boldsymbol{\Theta}_{\boldsymbol{\pi}_1}; \cdots; \boldsymbol{\Theta}_{\boldsymbol{\pi}_{G+2}}] \tag{26}$$

where $\boldsymbol{\pi}_i$ represents the random permutation, as shown in Section 4. The modified CLPSO is denoted as CLPSO-LDSEDS.

## 6.1 Test Functions

The test functions were still provided by the CEC 2017 special session [48]. As the search capability of CLPSO is better than that of PSO, we can select 19 test functions, listed in Table V, for which CLPSO can find optimal solutions with relative errors smaller than $\varepsilon_{\text{tol}} = 5\%$. The other 11 test functions are discarded.

TABLE VII

BENCHMARK FUNCTIONS

| Topology | No. | Function name | $Z^*$ |
|---|---|---|---|
| Unimodal Fns. | $F_1$ | Shifted and rotated Zakharov function | 300 |
| Simple multimodal Fns. | $F_2$ | Shifted and rotated Rosenbrock's function | 400 |
| | $F_3$ | Shifted and rotated Rastrigin's function | 500 |
| | $F_4$ | Shifted and rotated Expaned Schaffer F6 function | 600 |
| | $F_5$ | Shifted and rotated Lunacek Bi-Ratrigin's function | 700 |
| | $F_6$ | Shifted and rotated Non-continuous Rastrigin's function | 800 |
| | $F_7$ | Shifted and rotated Lvey function | 900 |
| Hybrid Fns. | $F_8$ | Zakharov, Rosenbrock, Rastrigin | 1100 |
| | $F_9$ | Bent Cigar, Rosenbrock; Lunacek bi-Rastrigin | 1300 |
| | $F_{10}$ | High-conditioned elliptic; Ackley; Schaffer; Rastrigin | 1400 |
| | $F_{11}$ | Bent Cigar; HGBat; Rastrigin; Rosenbrock | 1500 |
| | $F_{12}$ | Expanded schaffer; HGBat; Rosenbrock; Modified Schwefel; Rastrigin | 1600 |
| | $F_{13}$ | Katsuura; Ackely; Expanded Griewank plus Rosenbrock; Schwefel; Rastrigin | 1700 |
| | $F_{14}$ | High-conditioned elliptic; Ackley; Rastrigin; HGBat; Dicus | 1900 |
| | $F_{15}$ | Bent Cigar; Griewank plus Rosenbrock; Rastrigin; Expanded Schaffer | 2000 |
| Composite Fns. | $F_{16}$ | Katsuura; Ackley; Rastrigin; Schaffer; Modified Schwefel | 2100 |
| | $F_{17}$ | Griewank; Rastrigin; Modified schwefel | 2200 |
| | $F_{18}$ | Ackley; Griewank; Rastrigin; High-conditioned Elliptic | 2400 |
| | $F_{19}$ | Modified Schwefel; Rastrigin; Rosenbrock; Griewank; Expanded schaffer | 2600 |

## 6.2 Parameters Setting

The CLPSO code is downloaded from https://github.com/ zmdsn/ALFRAME. The inertia weight $w$ decreases linearly with iterations in the range of 0.9-0.2. The acceleration coefficient is $c = 1.49445$ . The maximum number of iterations is $G = 7500$ . When $D = 10$ , $N = 40$ , and $\varepsilon_{\text{tol}} = 1\%$ and $5\%$ were used. When $D = 30$ , $N = 100$, and $\varepsilon_{\text{tol}} = 10\%$ and 20% were used.

## 6.3 Numerical Comparisons and Discussions

A comparison between CLPSO and CLPSO-LDSEDS with 10 dimensional test functions is first presented. HWS, DES, HSS and SS were adopted to generate $\boldsymbol{\Theta}$ . The CSs, the CTs and their ranks of the results obtained by using the original CLPSO (denoted by Rand) and the CLPSO with different LDSs. (denoted by the name of the sampling method) are listed in Table VIII and Table IX.

As shown in Table VIII and Table IX, $\tau_F$ is greater than the critical value $\tau_c = 2.537$ for both $\varepsilon_{\text{tol}} = 5\%$ and $\varepsilon_{\text{tol}} = 1\%$ , which means that there are significant differences between the performances of all five algorithms considered. According to the NT, $CD = 1.399$ , the AvgRks of the CLPSO-LDSEDSs with HWS, DES, and SS are significantly smaller than those of CLPSO, which means that these three algorithms converge significantly faster than CLSPO at the same error tolerance. The AvgRk of CLPSO-LDSEDS with HSS is also smaller than that of the original CLPSO.

TABLE VIII

CSs AND RANKS OF THE FIVE ALGORITHMS AT DIFFERENT TOLERANCE ERRORS

| | $\varepsilon_{\text{tol}} = 5\%$ | | | | | $\varepsilon_{\text{tol}} = 1\%$ | | | | |
|---|---|---|---|---|---|---|---|---|---|---|
| | Rand | HWS | DES | HSS | SS | Rand | HWS | DES | HSS | SS |
| $F_1$ | 2285(5) | 2243(3) | 2203(2) | 2272(4) | 2116(1) | 2603(5) | 2602(4) | 2527(2) | 2589(3) | 2405(1) |
| $F_2$ | 437(5) | 418(3) | 409(2) | 433(4) | 385(1) | 2850(5) | 2755(2) | 2774(4) | 2770(3) | 2275(1) |
| $F_3$ | 972(5) | 857(2) | 876(3) | 956(4) | 785(1) | 3732(3.5) | 3761(5) | 3566(1) | 3732(3.5) | 3655(2) |
| $F_4$ | 127(5) | 111(1.5) | 120(3) | 122(4) | 111(1.5) | 510(5) | 442(3) | 438(2) | 509(4) | 427(1) |
| $F_5$ | 1568(5) | 1461(3) | 1409(2) | 1566(4) | 1252(1) | -(3) | -(3) | -(3) | -(3) | -(3) |
| $F_6$ | 387(3) | 385(2) | 400(4) | 462(5) | 349(1) | 2898(5) | 2728(3) | 2812(3) | 2869(4) | 2597(1) |
| $F_7$ | 549(4) | 527(3) | 519(2) | 561(5) | 487(1) | 846(4) | 785(2) | 764(2) | 896(5) | 729(1) |
| $F_8$ | 468(5) | 446(2) | 455(4) | 452(3) | 411(1) | 1180(5) | 1149(3) | 1110(2) | 1167(4) | 1086(1) |
| $F_9$ | 6898(5) | 4419(3) | 5961(4) | 4376(2) | 4137(1) | -(3) | -(3) | -(3) | -(3) | -(3) |
| $F_{10}$ | 2423(4) | 1925(1) | 2065(2) | 2215(3) | 2755(5) | -(4.5) | 7499(2.5) | 7499(2.5) | 7196(1) | -(4.5) |
| $F_{11}$ | 2254(5) | 2042(3) | 1829(1) | 2162(4) | 2030(2) | 6976(5) | 5874(3) | 5076(1) | 5597(2) | 6703(4) |
| $F_{12}$ | 601(5) | 506(1) | 591(3) | 597(4) | 524(2) | 1128(5) | 1024(1) | 1118(4) | 1097(3) | 1078(2) |
| $F_{13}$ | 315(4) | 306(3) | 268(1) | 344(5) | 274(2) | 2416(5) | 2373(3) | 2324(2) | 2401(4) | 2307(1) |
| $F_{14}$ | 1246(5) | 864(1) | 950(2) | 1221(4) | 1023(3) | 3294(5) | 2494(1) | 2806(3) | 2928(4) | 2775(2) |
| $F_{15}$ | 252(5) | 238(3) | 235(2) | 245(4) | 198(1) | 1330(5) | 1110(1.5) | 1177(3) | 1299(4) | 1110(1.5) |
| $F_{16}$ | -(4) | -(4) | 6884(2) | -(4) | 3407(1) | -(3) | -(3) | -(3) | -(3) | -(3) |
| $F_{17}$ | 683(5) | 677(4) | 584(1) | 627(2) | 641(3) | -(3) | -(3) | -(3) | -(3) | -(3) |
| $F_{18}$ | 4931(4) | 4634(2) | -(5) | 4746(3) | 3072(1) | -(3) | -(3) | -(3) | -(3) | -(3) |
| $F_{19}$ | -(4) | -(4) | 5858(2) | 5811(1) | -(4) | -(3) | -(3) | -(3) | -(3) | -(3) |
| AvgR. | 4.58(5) | 2.55(3) | 2.47(2) | 3.63(4) | 1.76(1) | 4.21(5) | 2.74(3) | 2.61(2) | 3.29(4) | 2.16(1) |
| $\tau_F$ | | | 17.287 | | | | | 5.947 | | |

TABLE IX

CTs and Ranks of the Five Algorithms at Different Tolerance Errors

| | $\varepsilon_{\text{tol}} = 5\%$ | | | | | $\varepsilon_{\text{tol}} = 1\%$ | | | | |
|---|---|---|---|---|---|---|---|---|---|---|
| | Rand | HWS | DES | HSS | SS | Rand | HWS | DES | HSS | SS |
| $F_1$ | 1.029(5) | 0.807(1) | 0.914(3) | 0.922(4) | 0.832(2) | 1.152(5) | 0.967(1) | 1.090(3) | 1.097(4) | 0.998(2) |
| $F_2$ | 0.286(5) | 0.222(1) | 0.284(4) | 0.280(3) | 0.248(2) | 1.246(4) | 1.146(1) | 1.239(3) | 1.284(5) | 1.226(2) |
| $F_3$ | 0.445(5) | 0.314(1) | 0.404(3) | 0.422(4) | 0.362(2) | 1.706(5) | 1.355(1) | 1.641(4) | 1.617(3) | 1.512(2) |
| $F_4$ | 0.193(4) | 0.165(2) | 0.181(3) | 0.205(5) | 0.157(1) | 0.227(5) | 0.165(1) | 0.196(3) | 0.221(4) | 0.191(2) |
| $F_5$ | 0.670(5) | 0.575(2) | 0.652(3) | 0.679(4) | 0.552(1) | 3.7680(5) | 3.184(1) | 3.702(3) | 3.736(4) | 3.677(2) |
| $F_6$ | 0.215(5) | 0.144(1) | 0.204(4) | 0.203(3) | 0.188(2) | 1.320(5) | 1.108(1) | 1.275(3) | 1.282(4) | 1.269(2) |
| $F_7$ | 0.234(5) | 0.178(1) | 0.215(3) | 0.228(4) | 0.194(2) | 0.413(4) | 0.316(1) | 0.386(3) | 0.419(5) | 0.349(2) |
| $F_8$ | 0.177(5) | 0.144(1) | 0.172(3) | 0.173(4) | 0.161(2) | 0.516(4) | 0.415(1) | 0.462(3) | 0.534(5) | 0.442(2) |
| $F_9$ | 1.795(5) | 1.294(1) | 1.713(3) | 1.756(4) | 1.575(2) | 3.257(5) | 2.698(1) | 3.235(4) | 3.109(2) | 3.139(3) |
| $F_{10}$ | 0.868(5) | 0.633(1) | 0.858(4) | 0.822(2) | 0.822(3) | 2.784(5) | 2.123(1) | 2.574(2) | 2.757(4) | 2.700(3) |
| $F_{11}$ | 0.932(5) | 0.631(1) | 0.695(2) | 0.865(4) | 0.786(3) | 2.153(5) | 1.796(1) | 2.013(2) | 2.035(3) | 2.044(4) |
| $F_{12}$ | 0.234(5) | 0.157(1) | 0.226(4) | 0.214(3) | 0.196(2) | 0.426(4) | 0.333(1) | 0.403(2) | 0.430(5) | 0.419(3) |
| $F_{13}$ | 0.144(5) | 0.108(1) | 0.140(3) | 0.143(4) | 0.130(2) | 1.153(5) | 0.962(1) | 1.057(3) | 1.123(4) | 1.030(2) |
| $F_{14}$ | 0.687(5) | 0.439(1) | 0.623(3) | 0.652(4) | 0.614(2) | 2.330(5) | 1.701(1) | 1.903(2) | 2.065(3) | 2.182(4) |
| $F_{15}$ | 0.098(5) | 0.073(1) | 0.081(2) | 0.087(4) | 0.087(3) | 0.585(5) | 0.439(1) | 0.522(2) | 0.579(4) | 0.543(3) |
| $F_{16}$ | 2.540(3) | 2.817(5) | 2.258(2) | 2.621(4) | 1.804(1) | 3.930(4) | 3.382(1) | 3.881(3) | 3.864(2) | 3.964(5) |
| $F_{17}$ | 0.423(5) | 0.328(1) | 0.362(2) | 0.398(4) | 0.387(3) | 4.311(3) | 3.781(1) | 4.240(2) | 4.344(5) | 4.333(4) |
| $F_{18}$ | 2.477(4) | 2.314(2) | 3.129(5) | 2.393(3) | 1.424(1) | 4.353(5) | 3.747(1) | 4.352(4) | 4.261(2) | 4.348(3) |
| $F_{19}$ | 3.095(5) | 2.878(2) | 3.054(3) | 2.659(1) | 3.078(4) | 4.660(5) | 4.004(1) | 4.510(3) | 4.514(4) | 4.421(2) |
| AvgR. | 4.790(5) | 1.421(1) | 3.105(3) | 3.579(4) | 2.105(2) | 4.632(5) | 1.000(1) | 2.842(3) | 3.790(4) | 2.737(2) |
| $\tau_F$ | | | 39.000 | | | | | 50.689 | | |

Next, the 30-dimensional test functions were used to test the performance of CLPSO-LDSEDS. In addition, some functions in Table V were discarded, as CLPSO cannot find their optimal solutions with relative errors smaller than 20%. The CSs , the CTs and their ranks of different algorithms are compared in Table X and XI. It can be observed again that CLPSO-LDSEDSs perform better than CLSPO. The Friedman statistics $\tau_F$ are larger than the critical value $\tau_c = 2.537$ for both the CSs and the CTs, which shows the significant differences between the considered five algorithms. According to the NT, when $\varepsilon_{\text{tol}} = 20\%$, the AvgRks of CSs of all four CLPSO-LDSEDSs are significantly smaller than that of CLPSO; when $\varepsilon_{tol} = 10\%$ , the AvgRks of CSs of CLPSO-LDSEDSs with HWS and DES are significantly smaller than that of CLPSO. The AvgRks of CTs of with HWS and DES are significantly smaller than that of CLPSO; when $\varepsilon_{tol} = 10\%$ , The AvgRks of CTs of CLPSO-LDSEDSs with HWS, DES, and SS are significantly smaller than that of CLPSO.

According to Tables VI-VII, it can be concluded that the proposed LDSEDS technique can substantially improve the convergence speed without reducing the search ability of the original CLPSO. Among the four adopted LDSs, the HWS and DES are more robust than the other two samplings and can significantly improve the convergence speed of CLPSO.

TABLE X

CSs AND RANKS OF THE FIVE ALGORITHMS AT DIFFERENT TOLERANCE ERRORS

| | $\varepsilon_{tol} = 20\%$ | | | | | $\varepsilon_{tol} = 10\%$ | | | | |
|---|---|---|---|---|---|---|---|---|---|---|
| | Rand | HWS | DES | HSS | SS | Rand | HWS | DES | HSS | SS |
| $F_2$ | 4086(5) | 3568(1) | 3700(2) | 3719(3) | 3827(4) | -(3) | -(3) | -(3) | -(3) | -(3) |
| $F_3$ | 2906(5) | 2567(1) | 2611(2) | 2801(3) | 2854(4) | 6139(5) | 5800(2) | 5619(1) | 5951(3) | 6084(4) |
| $F_4$ | 28(5) | 19(2) | 2(1) | 25(3) | 27(4) | 335(5) | 267(2) | 261(1) | 294(3) | 312(4) |
| $F_5$ | 3338(5) | 2693(1) | 2868(2) | 3061(3) | 3254(4) | -(3) | -(3) | -(3) | -(3) | -(3) |
| $F_6$ | 1633(5) | 1364(1) | 1403(2) | 1426(3) | 1595(4) | 4044(5) | 3571(1) | 3578(2) | 3671(3) | 3825(4) |
| $F_7$ | 3715(5) | 3079(1) | 3237(2) | 3560(3) | 3687(4) | 4454(5) | 3727(1) | 3855(2) | 4176(3) | 4373(4) |
| $F_8$ | 2195(5) | 1943(2) | 1933(1) | 2020(3) | 2091(4) | 3749(5) | 3434(2) | 3361(1) | 3604(3) | 3670(4) |
| $F_{11}$ | 4371(5) | 4072(1) | 4200(2) | 4251(3) | 4291(4) | 5845(5) | 5479(2) | 5776(4) | 5448(1) | 5634(3) |
| $F_{13}$ | 1179(5) | 1074(2) | 1081(3) | 1056(1) | 1155(4) | 2406(5) | 2137(3) | 2131(2) | 2116(1) | 2402(4) |
| $F_{14}$ | 3434(5) | 2990(1) | 3087(2) | 3145(3) | 3350(4) | 3922(5) | 3491(1) | 3623(3) | 3584(2) | 3827(4) |
| $F_{15}$ | 1279(5) | 959(1) | 1091(3) | 1162(4) | 1070(2) | 3578(4) | 2605(1) | 3542(4) | 7270(5) | 2810(2) |
| $F_{16}$ | 735(5) | 655(1) | 656(2) | 673(4) | 657(3) | -(4.5) | 5321(2) | -(4.5) | 6380(3) | 4162(1) |
| $F_{17}$ | 1938(5) | 1910(4) | 1630(2) | 1561(1) | 1753(3) | 3333(3) | 6182(5) | 2718(2) | 2609(1) | 3421(4) |
| $F_{18}$ | 4986(3) | 7310(5) | 4553(1) | 6482(4) | 4643(2) | -(3) | -(3) | -(3) | -(3) | -(3) |
| $F_{19}$ | -(3.5) | 6262(1) | -(3.5) | -(3.5) | -(3.5) | -(3) | -(3) | -(3) | -(3) | -(3) |
| AvgR. | 4.767(5) | 1.667(1) | 2.033(2) | 2.967(4) | 2.533(3) | 4.233(5) | 2.267(1) | 2.500(2) | 2.667(3) | 3.333(4) |
| $\tau_F$ | | | 22.416 | | | | | 4.817 | | |

TABLE XI

CTs AND RANKS OF THE FIVE ALGORITHMS AT DIFFERENT TOLERANCE ERRORS

| | $\varepsilon_{tol} = 20\%$ | | | | | $\varepsilon_{tol} = 10\%$ | | | | |
|---|---|---|---|---|---|---|---|---|---|---|
| | Rand | HWS | DES | HSS | SS | Rand | HWS | DES | HSS | SS |
| $F_2$ | 6.872(5) | 4.621(1) | 5.753(3) | 5.627(2) | 6.491(4) | 10.030(5) | 8.381(1) | 9.844(2) | 9.640(4) | 9.724(3) |
| $F_3$ | 4.454(5) | 3.263(1) | 3.933(2) | 4.114(3) | 4.262(4) | 10.067(5) | 8.237(1) | 9.543(2) | 9.025(3) | 9.661(4) |
| $F_4$ | 0.027(3) | 0.019(1) | 0.019(2) | 0.0290(5) | 0.028(4) | 0.522(5) | 0.409(1) | 0.461(3) | 0.486(2) | 0.498(4) |
| $F_5$ | 6.023(5) | 4.219(1) | 4.959(2) | 5.378(3) | 5.747(4) | 13.713(5) | 11.910(1) | 13.374(2) | 13.300(4) | 13.319(3) |
| $F_6$ | 2.351(5) | 1.638(1) | 1.970(2) | 2.085(4) | 2.043(3) | 6.224(5) | 4.825(1) | 5.628(3) | 5.791(2) | 5.841(4) |
| $F_7$ | 5.163(5) | 3.651(1) | 4.484(2) | 4.769(3) | 4.973(4) | 6.589(5) | 4.963(1) | 5.886(3) | 6.276(2) | 6.514(4) |
| $F_8$ | 2.600(5) | 1.962(1) | 2.435(2) | 2.467(3) | 2.527(4) | 5.127(5) | 4.180(1) | 4.629(3) | 4.885(2) | 5.102(4) |
| $F_{11}$ | 6.391(5) | 4.914(1) | 5.998(2) | 6.364(4) | 6.162(3) | 8.559(5) | 6.553(1) | 7.573(4) | 8.353(2) | 8.118(3) |
| $F_{13}$ | 1.803(5) | 1.544(1) | 1.702(2) | 1.783(4) | 1.780(3) | 4.631(5) | 3.436(2) | 3.303(3) | 3.926(1) | 4.162(4) |
| $F_{14}$ | 11.429(5) | 9.370(1) | 9.894(2) | 10.597(3) | 10.640(4) | 13.412(5) | 11.339(1) | 11.800(3) | 12.429(2) | 13.189(4) |
| $F_{15}$ | 2.431(5) | 1.640(1) | 2.297(3) | 2.341(4) | 1.9610(2) | 7.141(3) | 4.166(1) | 8.088(5) | 10.861(4) | 5.849(2) |
| $F_{16}$ | 1.186(5) | 0.969(1) | 1.168(4) | 1.126(3) | 1.055(2) | 13.401(4) | 12.222(3) | 15.839(2) | 11.733(5) | 10.591(1) |
| $F_{17}$ | 3.572(5) | 3.190(4) | 2.506(1) | 2.831(2) | 3.103(3) | 5.295(4) | 5.971(5) | 4.082(2) | 4.801(1) | 5.284(3) |
| $F_{18}$ | 15.678(2) | 15.812(3) | 15.388(1) | 17.482(4) | 17.598(5) | 17.609(2) | 16.213(1) | 17.623(5) | 17.892(3) | 17.823(4) |
| $F_{19}$ | 21.539(4) | 18.564(2) | 18.008(1) | 21.411(3) | 21.916(5) | 22.149(5) | 20.674(1) | 22.100(3) | 22.096(4) | 22.057(2) |
| AvgR. | 4.600(5) | 1.400(1) | 2.067(2) | 3.333(3) | 3.600(4) | 4.533(5) | 1.467(1) | 2.733(2) | 3.000(3) | 3.268(4) |
| $\tau_F$ | | | 25.573 | | | | | 13.155 | | |

# 7. Conclusions

Many researchers have tried to use LDS to initialize the population, hoping to accelerate the convergence of PSO by taking advantage of the excellent uniformity of the low-discrepancy sample set. However, the effect of this improvement direction seems inconclusive, and it also lacks corresponding theoretical support.

Based on Niderreiter's theorem, this paper reveals the relationship between the uniformity of the sample set and the error bound of PSO, which provides theoretical support for using LDS in the expanded dimensional space to accelerate PSO. On this basis, a new acceleration technique, LDSEDS, is provided. Two different sample construction methods are proposed to generate the low-discrepancy sample sets in the expanded dimensional space, and correspondingly, two more effective PSO algorithms are proposed, namely PSO-LDSEDS1 and PSO-LDSEDS2.

The computational performance of PSO-LDSEDS1 and PSO-LDSEDS2 is studied, and the following conclusions are obtained based on numerical verification. Using PSO-LDSEDS1 to calculate the test functions can indeed converge faster than PSO. However, this advantage is only obvious when the population size $N$ is large due to the construction method of the low-discrepancy sample set in PSO-LDSEDS1. Therefore, we prefer to use PSO-LDSEDS2, which has the same advantages and requires fewer samples. Numerical experiments show that the proposed PSO-LDSEDS2 not only improves the convergence speed of PSO in solving low dimensional problems but also improves its convergence speed in solving high-dimensional problems.

In addition, the improved direction asserted in this paper can be extended to other PSO-type algorithms. The CLPSO-LDSEDS generated by combining with CLPSO also has a fast convergence speed in low and high dimensions.

The effects of the different LDSs are different but on the whole, HWS and DES are more robust. There is always a significant difference between the improved algorithm using HWS (or DES) and the original algorithm. Therefore, it is recommended to give priority to HWS and DES in practical application.

Finally, it is worth further emphasizing that the acceleration method in this paper may be extended to other PSO-type optimization algorithms (including multiobjective optimization problems) and to other types of evolutionary algorithms, which is also a direction needing further study.

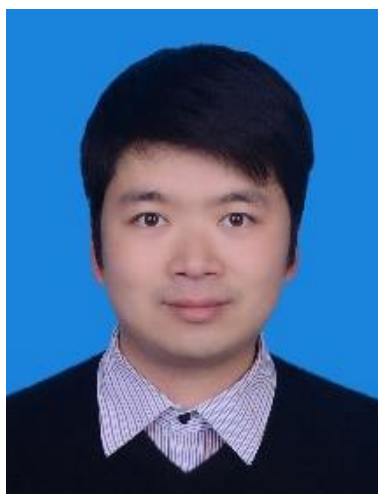

**Feng Wu** (*Member, IEEE*) received his B.Sc degrees in Civil Engineering from Guangxi University in 2007. And from the same university, he received the M.Eng in Structural Engineering in 2010. He obtained the Ph.D degree in Engineering Mechanics from Dalian University of Technology in 2015. From 2015 to 2017, he worked as a post-doctor in the School of Naval Architecture & Ocean Engineering at Dalian University of Technology. Since 2017, he has been an associate professor in Dalian University of Technology. His research direction is the dynamics and control.

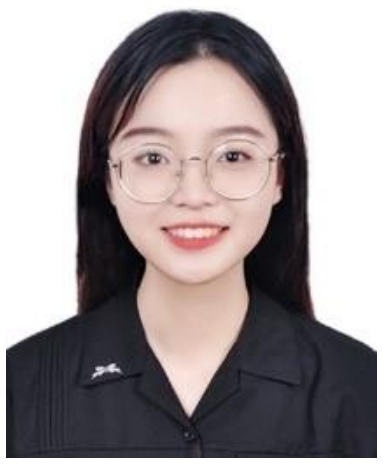

**Yuelin Zhao** (*Student Member*, IEEE) is working as a postgraduate student in Computational mechanics with the faculty of vehicle engineering and mechanics, Dalian University of Technology. Her research interests include the construction and application of the low-discrepancy samples.

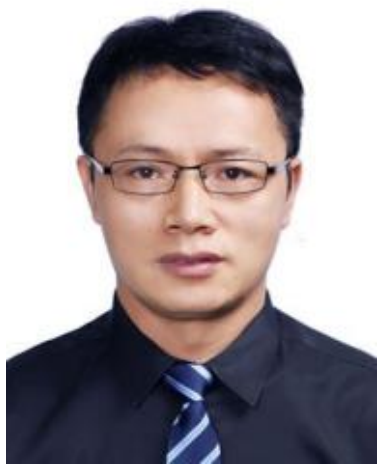

**Jianhua Pang** received the Ph.D degree in Ship and Ocean Engineering from Dalian University of Technology in 2016. Currently, he is the leader of the marine intelligent equipment and system team of Guangdong Ocean University Shenzhen Research Institute, the post doctoral supervisor of marine engineering technology of Tsinghua University International Graduate School, the young scientist of Guangdong Province, the outstanding young scholar of Guangdong Ocean University's top talents, the high-level leading reserve talents of Shenzhen. His research direction is the theory and method of intelligent ocean technology.

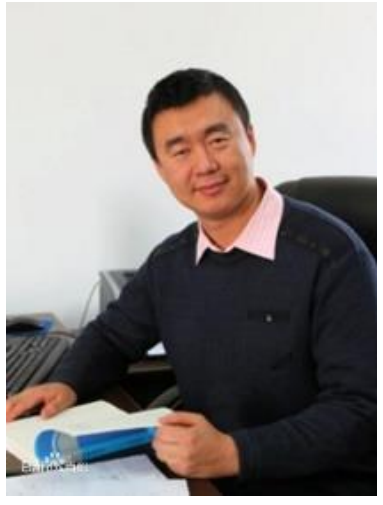

**Jun Yan** received the Ph.D degree in Engineering Mechanics from Dalian University of Technology in 2007. In 2015, he was selected into the growth plan of outstanding young scholars in colleges and universities of Liaoning Province. He is the deputy director of the Department of Engineering Mechanics of Dalian University of Technology, the member of the Education Work Committee of the Chinese Mechanics Society, the member of the Foreign Exchange and Cooperation Work Committee, and the director of key funds of the National Natural Science Foundation of China. His research direction is structure and multidisciplinary optimization.

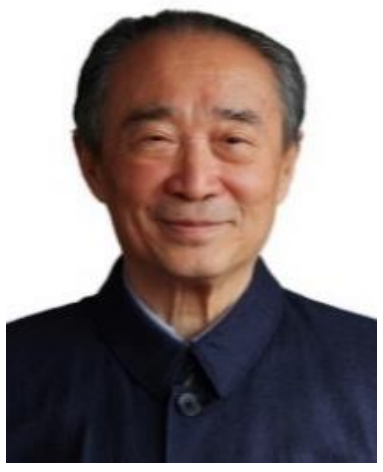

**Wanxie Zhong** graduated from the Department of Bridge and Tunnel, Tongji University in 1956. In 1984, he was elected as the chairman of China Association of Computational Mechanics; In 1993, he was awarded honorary professor by the University of Wales and the University of Hong Kong; In the same year, he was elected as an academician of Chinese Academy of Sciences. He has put forward important theories and methods in group theory, limit analysis, parametric variational principle, etc. , and organizes and develops various large-scale structural simulation systems.